\documentclass[conference]{IEEEtran}
\IEEEoverridecommandlockouts

\usepackage{amsmath,amssymb,amsfonts}
\usepackage{hyperref}
\usepackage{algorithmic}
\usepackage{graphicx}
\usepackage{float}
\usepackage{textcomp}
\usepackage{booktabs}
\usepackage{multirow}
\usepackage{listings}
\usepackage{xcolor}
\usepackage{tabularx} 
\usepackage{lipsum} 
\usepackage[acronym]{glossaries}
\usepackage[sorting=none, style=numeric]{biblatex} 
\glsdisablehyper
\newacronym{AI}{AI}{Artificial Intelligence}
\newacronym{ANN}{ANN}{Artificial Neural Network}
\newacronym{CNN}{CNN}{Convolutional Neural Network}
\newacronym{CPU}{CPU}{Central Processing Unit}
\newacronym{CSNN}{CSNN}{Convolutional Spiking Neural Network}
\newacronym{DL}{DL}{Deep Learning}
\newacronym{DNN}{DNN}{Deep Neural Network}
\newacronym{DVS}{DVS}{Dynamic Vision Sensor}
\newacronym{FC}{FC}{Fully Connected}
\newacronym{FF}{FF}{Feed Forward}
\newacronym{FI}{FI}{Fault Injector}
\newacronym{FIM}{FIM}{Fault Injection Manager}
\newacronym{FL}{FL}{Fault List}
\newacronym{FLG}{FLG}{Fault List Generator}
\newacronym{FPGA}{FPGA}{Field Programmable Gate Array}
\newacronym{LIF}{LIF}{Leaky Integreted and Fire}
\newacronym{NLP}{NLP}{Natural Language Processing}
\newacronym{NN}{NN}{Neural Network}
\newacronym{SDC}{SDC}{Silent Data Corruption}
\newacronym{SFI}{SFI}{Statistical Fault Injection}
\newacronym{SHD}{SHD}{Spiking Heildelberg Digits}
\newacronym{SNN}{SNN}{Spiking Neural Network}
\newacronym{SoA}{SoA}{State-of-Art}
\newacronym{STDP}{STDP}{Spike Timing Dependent Plasticity}

\def\BibTeX{{\rm B\kern-.05em{\sc i\kern-.025em b}\kern-.08em
    T\kern-.1667em\lower.7ex\hbox{E}\kern-.125emX}}

\begin{document}

\newcommand{\spikingjet}{SpikingJET}

\title{SpikingJET: Enhancing Fault Injection for Fully and Convolutional Spiking Neural Networks
\thanks{Identify applicable funding agency here. If none, delete this.}
}

\author{\IEEEauthorblockN{Anıl Bayram Göğebakan, Enrico Magliano, Alessio Carpegna, Annachiara Ruospo, \\ Alessandro Savino, Stefano Di Carlo}
\IEEEauthorblockA{\textit{Control and Computer Department} \\
\textit{Politecnico di Torino}\\
Torino, Italy \\
Contact: stefano.dicarlo@polito.it}
}

\maketitle

\begin{abstract}
As artificial neural networks become increasingly integrated into safety-critical systems such as autonomous vehicles, devices for medical diagnosis, and industrial automation, ensuring their reliability in the face of random hardware faults becomes paramount. This paper introduces \spikingjet{}, a novel fault injector designed specifically for fully connected and convolutional \glspl{SNN}. Our work underscores the critical need to evaluate the resilience of SNNs to hardware faults, considering their growing prominence in real-world applications. \spikingjet{} provides a comprehensive platform for assessing the resilience of \glspl{SNN} by inducing errors and injecting faults into critical components such as synaptic weights, neuron model parameters, internal states, and activation functions. This paper demonstrates the effectiveness of \spikingjet{} through extensive software-level experiments on various \gls{SNN} architectures, revealing insights into their vulnerability and resilience to hardware faults. Moreover, highlighting the importance of fault resilience in \glspl{SNN} contributes to the ongoing effort to enhance the reliability and safety of \gls{NN}-powered systems in diverse domains.
\end{abstract}


\glsresetall

\section{Introduction}
\label{sec:introduction}


\gls{AI}, and in particular the field of \gls{DL} and \glspl{DNN}, is gradually changing the research and industrial landscape, allowing to solve complex problems with unprecedented precision, reaching human capabilities in areas such as image classification \cite{russakovsky}. This makes it particularly attractive in various fields, including safety-critical applications like avionics, automotive, robotics, and healthcare.


\gls{SNN} emerge as an energy-efficient solution for resource and power-constrained applications in this landscape~\cite{Carpegna:2022,carpegna2022artificial}. They take direct inspiration from neuroscience observations, operating on asynchronous events resembling the action potentials, or spikes, used by biological neurons to communicate. Consequently, they excel at the real-time processing of dynamic, time-sensitive data in edge-computing scenarios. 


Nonetheless, to deploy \glspl{SNN} in safety-critical systems, it is crucial to assess their reliability since a wrong decision can harm individuals or the system itself. One of the most commonly used and effective approaches to evaluate the robustness of \gls{DNN}-based applications is to perform extensive fault-injection campaigns, introducing faults in the system during its inference operation, gathering results, and comparing them with a fault-free execution, to assess the impact of the faults on the system's performance. 
%
%
While there are previous investigations into the overall effects of faults on individual spiking neurons\cite{spyrou_neuron_2021}, as well as comprehensive studies on full \glspl{SNN} \cite{nagarajan_analysis_2022, nagarajan_fault_2022}, they tend to investigate the reliability of specific neuron models, network architectures, or fault mechanisms. Furthermore, existing fault injectors \cite{colucci_enpheeph_2022}, although theoretically applicable to \glspl{SNN}, are primarily evaluated using traditional \gls{DNN} models. Hence, there is still the need for an extensive and flexible \gls{FI}, specifically tailored and validated for use with \glspl{SNN}, capable of assessing various fault types across different network architectures, neuron models, and applications.\\
%
%
This paper proposes \spikingjet{}, a simulation-based, non-intrusive \gls{FI} for \glspl{SNN}, built on top of the open-source \texttt{SnnTorch} \cite{eshraghian_training_2023} framework and developed as an extension of the tool proposed in \cite{gavarini_sci_fi_2023}, allowing the user to perform extensive campaigns, injecting stuck-at faults in static model parameters, such as synaptic weights, and neuron's internal parameters, taking into account specific \gls{SNN} features. Additionally, \spikingjet{} can inject faults into dynamic state variables and activation functions, ensuring the persistence of stuck-at faults throughout the inference process. An extensive study of fault injection on \glspl{SNN} through \spikingjet{} is presented to demonstrate the tool's effectiveness. The injection campaign targets three \gls{SNN} models, evaluated on as many benchmark datasets: N-MNIST~\cite{orchard_converting_2015}, \gls{SHD}~\cite{cramer_heidelberg_2022} and \gls{DVS} 128 gestures~\cite{amir_low_2017}, considering Convolutional,  \gls{FF} and recurrent \gls{FC} architectures. \\
%
%
The paper is structured as follows: \autoref{sec:related_works} reviews state-of-the-art solutions for fault tolerance in \glspl{SNN}. \autoref{sec:background} briefly overviews \gls{SNN} principles and fault injection techniques. \autoref{sec:proposed} outlines methodologies and design choices. \autoref{sec:experimental} details fault-injection campaigns with \spikingjet{} and discusses outcomes. Finally, \autoref{sec:conclusions} summarizes the study.
\section{Background}
\label{sec:background}


In the field of \glspl{ANN}, \glspl{SNN} stand out for their closer proximity to neuroscience models. First, neurons exchange information in the form of sequences of asynchronous spikes. The most common approach is to encode information with spike timing without focusing on their specific shape~\cite{auge_survey_2021}. In the digital realm, spikes are represented as single-bit binary events: either "present" (1) or "not present" (0). This simplifies complexity from spatial to temporal dimensions, reducing computational resources for static data like images. For temporally evolving data, appropriate encoding captures essential information, optimizing computations. This makes \glspl{SNN} particularly intriguing for resource-limited and power-constrained applications while capturing the temporal dynamics. Among the computational models developed in the last decades, this work targets the \gls{LIF} approach~\cite{burkitt_review_2006}, as it balances computational efficiency, temporal information capture, and biological fidelity. The discrete-time characteristic equation of the model is reported in \autoref{eq:lif}. 

    \begin{equation}
        \resizebox{0.9\columnwidth}{!}{%
        $V_m[n] = \begin{cases}
			\beta \cdot V_m[n-1] + W \cdot s_{in}[n], & \mbox{if } V_m[n-1] \leq V_{th} \\
            V_m[n-1] - V_{th} + W \cdot s_{in}[n], & \mbox{if } V_m[n-1] > V_{th}
        \end{cases}$
        }
		\label{eq:lif}
    \end{equation}

The model's core is $V_m$, which emulates the membrane potential of a biological neuron. To draw a parallel with artificial neurons, two subsequent operations can be identified: (i) a linear combination of the inputs, here expressed as a dot product between the synaptic weights $W$ and the input spikes $s_{in}[n]$ to identify a weighted sum of the inputs. For different models, like \gls{CSNN}, this can be a more complex operation, like a convolution. The result can be seen as the net input current received by the neuron. (ii) A non-linear function. Unlike traditional activation functions such as ReLU or sigmoid functions, the non-linearity is applied directly to the state variable $V_m$, which retains previous information ($V_m[n-1]$) and integrates the weighted sum. Finally, a threshold $V_{th}$ separates two working regimes: in sub-threshold, the membrane potential is iteratively decayed ($\beta \cdot V_m[n-1]$, $\beta < 1$), following an exponential trend. If instead $V_m$ exceeds the threshold, it is reset by decreasing it to a quantity equal to $V_{th}$. When this happens, an output spike $s_{out}$ is generated and sent to the subsequent units, as shown in \autoref{eq:fire}
 
        \begin{equation}
            s_{out}[n] = \begin{cases}
    			1, & \mbox{if } V_m[n-1] > V_{th} \\
    			0, & \mbox{if } V_m[n-1] \leq V_{th}
            \end{cases}
            \label{eq:fire}
        \end{equation}
		
Looking at the model from a fault-injection perspective, there are 5 possible points of fault: (i) the input linear operation involving synaptic weights $W$ and, if present, a bias value, (ii) the decay factor $\beta$, (iii) the threshold value $V_{th}$, (iv) the state variable $V_m$ and (v) the activation function $s_{out}$. $s_{in}$ is inherently considered as it represents the inputs originating from a preceding layer, constituting the ensemble of $s_{out}$ from that layer.

When assessing the resilience of a \gls{NN} model, the goal is to understand the impact of hardware faults on the model inference. This can be done by artificially injecting faults into the system and then comparing the outcome of the inference with the fault-free correct output, generally called a golden reference. Fault-injection can be performed at different levels of abstraction, namely at the system level, considering the hardware platform on which the \gls{NN} is executed, or at the application level, targeting a technology-independent model implementation. 
Faults in electronic devices can be classified based on their temporal nature: in transient bit-flips, an individual memory cell undergoes a local state change, which is correctly restored the first time the cell is written again. Conversely, permanent or stuck-at faults represent irreversible physical damage to the device, in which individual electronic elements are tied to a logical state (0 or 1)\cite{10.5555/2588043}.
Three primary methods are employed for fault injection ~\cite{Mamone:2020,portolan2019alternatives}. Simulation-based techniques involve modeling various fault mechanisms using software simulations or system emulation. Platform-based \gls{FI} involves manipulating the target hardware or software directly to induce faults. Lastly, radiation-based methods expose electronic devices to sources like alpha particles, neutrons, or electromagnetic radiation. As techniques approach the target hardware, faults are simulated more accurately, but complexity and control decrease. Therefore, simulation-level injection is ideal for broad, repeatable, and controllable assessment of system resiliency.

To address the increasing costs associated with fault injection procedures, \glspl{SFI} are today considered a valid solution to estimate a specific characteristic of a population of faults by observing only a reduced portion (i.e., a sample). This statistical measurement will introduce an error (i.e., error margin), meaning that the characteristic of the population being estimated will be close to the estimate by plus or minus the margin of error, which will happen with a specific confidence level. Hence, in a \gls{SFI}, the \gls{FL} includes a random sample of faults that is representative of the entire fault universe, and, given a specific error margin ($e$) and confidence level ($t$), the number ($n$) of injected fault is defined as in \autoref{eq:bernoullisampling}.

\begin{equation}\label{eq:bernoullisampling}
   n=\frac{N}{1+e^{2}\cdot\frac{N-1}{t^{2} \cdot p*(1-p)}}
\end{equation}

In \autoref{eq:bernoullisampling}, $N$ is the size of the total population of faults (this number depends on the adopted fault model), and $p$ is the probability of success. Adopting an equal probability of success or failure is a conservative solution. Hence, $p$ is equal to 0.5. 
It is worth underlining that \gls{SFI} techniques have been widely applied in the literature for estimating the vulnerability of \gls{DNN}s at different granularity (e.g., to find out the most critical layer, kernel, weight, or bit position). In \cite{DATE23}, the effectiveness of \glspl{SFI} in estimating failure rates (e.g., the number of wrong predictions caused by faults) has been experimentally validated with \glspl{CNN}. The statistical methodology shows that by injecting only the 1.21\% (ResNet-20) or 0.55\% (MobileNetV2) of the entire possible experiments, it is possible to achieve an estimate of the \gls{CNN} reliability close to the exhaustive result with an error always lower than 1\%.
\section{Related works}
\label{sec:related_works}

Fault injection in \glspl{SNN} can be approached in different ways \cite{stratigopoulos_testing_2023}. One common strategy is to model faults starting from typical hardware implementations of spiking neurons~\cite{el-sayed_spiking_2020}, trying to understand the possible faults that can incur and simulating them~\cite{spyrou_neuron_2021}. Functionally, these faults are typically classified into three categories: (i) dead neurons, (ii) saturated neurons (where the output spike consistently remains at 0 or 1, respectively, regardless of input), and (iii) timing variations (resulting from alterations in internal parameters affecting spike timing). The first two categories often stem from stuck-at faults, where the output spike or threshold becomes fixed at specific logical values, or faults in the membrane perpetually hold the neuron above or below the threshold, causing continuous firing or complete inhibition. Conversely, the third category encompasses other fault mechanisms, including milder stuck-at faults in the internal parameters or deviations in membrane behavior. Despite existing efforts, there remains a gap in addressing diverse fault types comprehensively within \glspl{SNN}.
Current injection methods often target specific hardware implementations or components, such as power supply \cite{nagarajan_analysis_2022, nagarajan_fault_2022}, or hardware memory \cite{wicaksana_putra_respawn_2021}, with limited focus on emerging hardware designs  \cite{vatajelu_special_2019}). Although some studies consider general resiliency to noise in spiking layers \cite{li_reliability_2023}, software-based \glspl{FI}, while allegedly compatible with \glspl{SNN}, primarily focus their analysis on more traditional models like \glspl{DNN}~\cite{colucci_enpheeph_2022}, leaving gaps in comprehensive fault assessment and resiliency testing across various architectures, neuron models, and applications.
\section{Proposed Approach}
\label{sec:proposed}



\begin{figure}[ht]
        \centering
            \includegraphics[width=0.98\columnwidth]{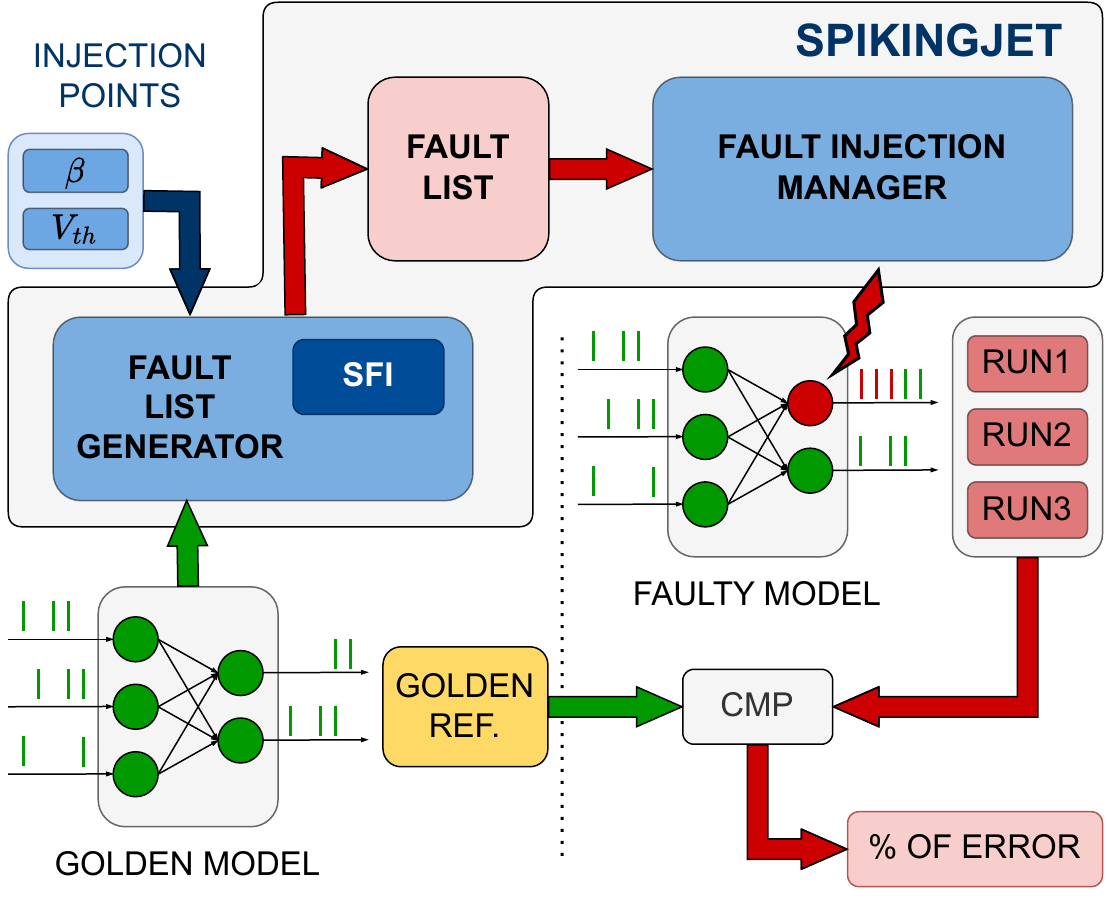}
        \caption{\spikingjet{} framework}
        \label{fig:spikingjet}
    \end{figure}

\autoref{fig:spikingjet} shows the main building blocks of the proposed \spikingjet{} framework. The tool consists of two primary components: the \gls{FLG}  and the \gls{FIM}, each playing a distinct role in the fault injection process. Specifically, the \gls{FLG} compiles the list of faults to be injected, enumerating them within the \gls{FL}. Subsequently, the \gls{FL} is passed to the \gls{FIM}, which utilizes it to execute the actual fault injection campaign. Finally, the results of the faulty runs are compared to the golden reference to assess the impact of each fault on the output accuracy. Let's delve deeper into the details of the process. It is crucial to note that the fault injection campaign occurs during the inference phase, necessitating the user's provision of a fully pre-trained model. Initially, this model (Golden Ref. in \autoref{fig:spikingjet} generates a fault-free reference, termed the golden reference, encompassing predictions for all input data.

%
%
The initial step involves \gls{FL} generation: the \gls{FLG} reads user requests (injection points in \autoref{fig:spikingjet}) and automatically retrieves all potential parameters available for injection within the network and verifies the compatibility of user requests with them. Subsequently, it constructs the complete \gls{FL}, containing the exhaustive list of single-bit faults to be injected within the \gls{SNN}. Here, the \gls{FIM} must interpret each entry of the \gls{FL} to ascertain the precise bit position for fault injection. Each entry specifies (i) the parameter for injection, which could be synaptic weights ($W$), neuron parameters ($V_{th}$, $\beta$), evolving state ($V_m$), or activation function ($s_{out}$), (ii) the spatial coordinates of the selected parameter within the network, indicating the specific layer, neuron, and connection for injection, and (iii) the index of the bit for injection.
Since an exhaustive fault injection in all the bits of all the requested parameters is generally unfeasible due to the size of the considered architectures, the \gls{FLG} adopts a stochastic approach based on the \gls{SFI} formula to ensure a thorough exploration when generating thee \gls{FL}, as explained in \autoref{sec:background}.

%
The second step consists of the effective fault injection campaign. The \gls{FIM} sequentially reads each entry of the \gls{FL} and executes a complete inference cycle for each, introducing the corresponding fault and recording the network's output predictions for subsequent comparison with the golden reference.
The persistence of the fault within the network throughout the entire inference process is ensured via two distinct methods, contingent on the target injection point:  if the target parameter is static, meaning it is read-only, it is corrupted before starting the inference process. Once corrupted, no further actions are necessary as the altered value remains unchanged during inference. This applies to synaptic weights, thresholds, and exponential decay constants. Conversely, the fault must be refreshed for dynamic parameters each time the value is rewritten. This is obtained by precisely instrumenting the code to access the target variable during the forward pass and to be able to actively modify it at runtime throughout the inference cycle. To simplify matters, the fault is refreshed at every iteration, regardless of the actual rewriting of the target parameter. This approach is adopted for both membrane potentials and output spikes.

\spikingjet{} supports the injection of stuck-at faults across all five points outlined in \autoref{sec:background}, encompassing three primary fault models:

\begin{enumerate}
    \item Stuck-at faults affecting synaptic connections
    \item Byzantine neurons, indicating timing variations in the firing activity of neurons' outputs
    \item Crashed neurons, specifically dead and saturated neurons
\end{enumerate}

The analysis aims to evaluate the sensitivity of any targeted location to faults, discerning the impact each has across various layers. Consequently, injection results are averaged across different bit positions of the fault occurring within the same parameter and neurons within the same layer.

\section{Experimental Results}
\label{sec:experimental}

\subsection{Experimental setup}
\label{subsec:exp_setup}

\spikingjet{} is evaluated on three distinct models, targeting three different datasets, obtained through specialized sensors or conversion chains, specifically developed to be used with \glspl{SNN}:

    \begin{enumerate}
        \item N-MNIST \cite{orchard_converting_2015}: spiking version of the original frame-based MNIST\cite{lecun_gradient-based_1998} dataset, obtained moving an ATIS \cite{posch_live_2010} event camera in front of the 28x28 8-bit greyscale images of handwritten digits. It counts 10 classes, with numbers from 0 to 9. A fully connected network with one hidden layer of 500 neurons and one output layer of 10 neurons is used to classify it, reaching a golden accuracy of 83.58\%.
        \item \gls{SHD} \cite{cramer_heidelberg_2022}: a dataset of spoken digits with 20 classes corresponding to numbers from 0 to 9 pronounced in English and German. Data were converted into trains of spikes using a biologically plausible model of the human cochlea. In this case, a fully connected recurrent model features a hidden layer with 200 neurons and an output layer of 20, with fully connected feedback connections between neurons in the hidden layer. The golden accuracy is 65.86\%.
        \item DVS128 \cite{amir_low_2017}: dataset of 11 gestures recorder through a \gls{DVS} camera\cite{lichtsteiner_128_2008}. In this case, a convolutional model is used with a 12C5-32C5-800-11 architecture, where $xCy$ identifies a convolutional layer with $x$ channels and a kernel of size $y \cdot y$. The golden accuracy is 72.26\%.
    \end{enumerate}


The selected models don't aim at reaching \gls{SoA} accuracy on the three benchmarks, which attest around 99.6\% for the N-MNIST \cite{samadzadeh_convolutional_2023}, 98.0\% for the DVS128 gesture \cite{she_sequence_2022} and 95.1\% for the \gls{SHD}\cite{hammouamri_learning_2024}, and their optimization is out of the scope of this work. The goal was to obtain reference models to evaluate the impact of the fault injection through \spikingjet{}.

\begin{table}[htb]
    \centering
    \caption{Breakdown of all potential target parameters and variables, aggregate per layer typology.}
    \label{tab:layer_param}
    \begin{tabular}{|c|c|c|}
        \hline
        \textbf{Layer}                      & 
        \textbf{Framework}                  &
        \textbf{Injectable Parameters}      \\
        
        \hline
        \textbf{Fully Connected}            &
        Torch                               &
        weight, bias                        \\

        \hline
        \textbf{Convolutional}              &
        Torch                               &
        weight, bias                        \\

        \hline
        \textbf{\gls{LIF}}                  &
        snnTorch                            &
        beta, threshold, potential, spike   \\
        \hline
        
    \end{tabular}
\end{table}

A first run is used by \spikingjet{} to extract the injectable parameters needed to build the \gls{FL}.\autoref{tab:layer_param} reports all the parameters and state variables targeted by the \gls{FI} per layer type, in line with what seen in \autoref{sec:background}, and they represent the Injection Point as depicted in \autoref{fig:spikingjet}. It is worth underlining that the adopted neural networks are all based on \emph{Float32} torch tensors. 
Fault injections have been performed with an error margin of $e=1\%$ and a confidence level $t=99\%$ (see \gls{SFI} in \autoref{eq:bernoullisampling}), following~\cite{date09_formula} . 
Different metrics are considered to classify all types of \gls{SDC}. In particular, the top-ranked element in the output scores vector, indicating the \gls{SNN} prediction. As such, the impact of the fault can be classified as:

    \begin{enumerate}
        \item Masked: top-ranked elements predicted scores are the same between faulty and golden runs.
        \item SDC-1: The top-ranked element predicted differs from that predicted by its error-free execution. This type of \gls{SDC} is particularly critical since it changes the network prediction.
        \item SDC min-max\%: The confidence score of the top-ranked element deviates between min\% and +/-max\% from its value in the error-free execution.
        
    \end{enumerate}

\subsection{Results}
\label{sec:results}



Table \ref{tab:ds_info} summarizes the injection campaigns for the three datasets, showing details about the amount of injected faults and the total time required for the injection.


\begin{table}[htb]
    \centering
    \caption{Total amount of injected fault and time required for a complete fault-injection campaign}
    \label{tab:ds_info}
    \begin{tabular}{|c|c|c|}
        \hline
        \textbf{Network Model}      & 
        \textbf{Injected faults}    &
        \textbf{FI time [h:m:s]}    \\
        
        \hline
        \textbf{DVS128}         &
        16,307                     &
        1:42:25                     \\

        \hline
        \textbf{N-MNIST}            &
        15,944                     &
        4:59:06                     \\

        \hline
        \textbf{SHD}                &
        16,578                    &
        1:09.44                     \\
        \hline
    \end{tabular}
\end{table}

The first thing to notice is that \spikingjet{} can perform a complete injection campaign involving all the available parameters within the network, with an overall time requirement between 1 and 5 hours, in line with the results reported in \cite{gavarini_sci_fi_2023}. This confirms that the original tool was extended with the capability to inject in \gls{SNN} typical parameters, maintaining similar time constraints and enabling a fast exploration of different fault mechanisms in \glspl{SNN}.

\begin{figure}[htb]
        \centering
            \includegraphics[width=0.98\columnwidth]{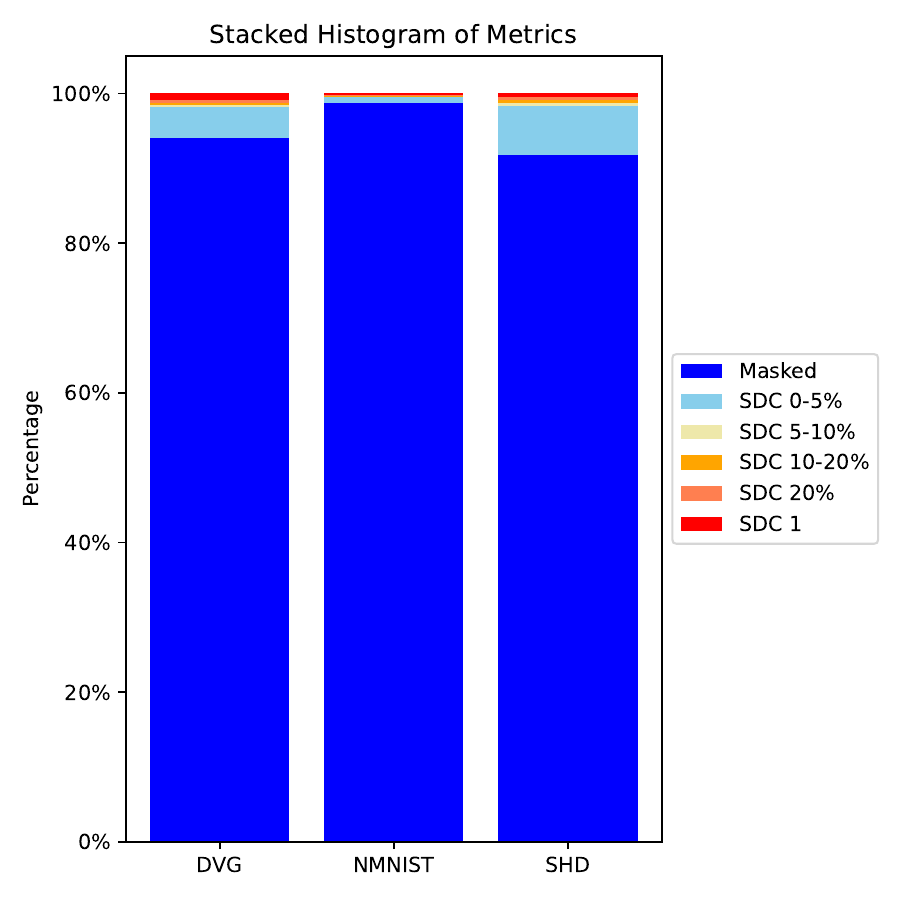}
        \caption{Overall classification results on the different benchmarks}
        \label{fig:net_agr}
    \end{figure}

The outcomes of the fault injection campaign, utilizing the specified parameters across the entire network for the three datasets, are depicted in \autoref{fig:net_agr}. Notably, most faults belong to the masked category, corroborating the inherent resilience of \glspl{NN} as reported in prior studies \cite{DATE23}. Furthermore, it appears that simpler networks exhibit greater resilience. However, the analysis conducted may not be exhaustive enough to establish this observation as a general rule: the \gls{FF}-\gls{FC} network used to classify the N-MNIST dataset exhibits only minor susceptibility to the injections, whereas the convolutional and recurrent models employed for the DVS128 and \gls{SHD} datasets display a higher degree of vulnerability. Moreover, among the unmasked faults, only a small subset results in complete mispredictions ($1\% \leq SDC-1 \leq 6.5\%$), and generally, the proportion of faults leading to an \gls{SDC} greater than 5\% is quite restricted.




For completeness, a more in-depth analysis is undertaken, examining vulnerability data at the granularity of individual layers. The findings are presented in \autoref{tab:faults_layer}. 
The percentage of faults injected relative to the total injectable parameters for each layer is reported as $n\%$. For instance, taking the second \gls{FC} layer, denoted as fc2 in the N-MNIST section of \autoref{tab:faults_layer}, it connects the $500$ neurons of the hidden layer with the $10$ neurons of the output layer, resulting in a total of $500 \cdot 10 = 5000$ weights. In this scenario, with $n\% = 3.75\%$, an average of $0.0375 \cdot 5000 = 188$ weights were injected with stuck-at faults.

\begin{table*}[t!]
 \caption{Analysis of the layer-wise impact of faults on the three benchmark datasets}
    \label{tab:faults_layer}
    \centering
\begin{tabular}{|cccccccccc|}
\hline
\multicolumn{10}{|c|}{NMNIST}                                                                                                                                                                  \\ \hline
\multicolumn{1}{|c|}{Layer Name}            & \multicolumn{1}{c|}{Layer Type}                       & \multicolumn{1}{c|}{Parameter} & \multicolumn{1}{c|}{n\%} & \multicolumn{1}{c|}{SDC 1} & \multicolumn{1}{c|}{SDC 0-5\%} & \multicolumn{1}{c|}{SDC 5-10\%} & \multicolumn{1}{c|}{SDC 10-20\%} & \multicolumn{1}{c|}{SDC 20\%} & Masked \\ \hline
\multicolumn{1}{|c|}{\multirow{2}{*}{fc1}}  & \multicolumn{1}{c|}{\multirow{2}{*}{Fully Connected}} & \multicolumn{1}{c|}{Bias}      & 4.84                     & 0.58                       & 5.58                           & 0.26                            & 0.30                             & 0.50                          & 92.75  \\
\multicolumn{1}{|c|}{}                      & \multicolumn{1}{c|}{}                                 & \multicolumn{1}{c|}{Weight}    & 4.18                     & 0.15                       & 0.67                           & 0.03                            & 0.03                             & 0.05                          & 99.04  \\ \hline
\multicolumn{1}{|c|}{\multirow{4}{*}{lif1}} & \multicolumn{1}{c|}{\multirow{4}{*}{Leaky}}           & \multicolumn{1}{c|}{Beta}      & 3.90                     & 0.69                       & 7.16                           & 0.30                            & 0.43                             & 0.67                          & 90.72  \\
\multicolumn{1}{|c|}{}                      & \multicolumn{1}{c|}{}                                 & \multicolumn{1}{c|}{Potential} & 4.37                     & 1.99                       & 7.03                           & 0.27                            & 0.43                             & 0.73                          & 89.51  \\
\multicolumn{1}{|c|}{}                      & \multicolumn{1}{c|}{}                                 & \multicolumn{1}{c|}{Spike}     & 4.21                     & 3.65                       & 10.89                          & 0.36                            & 0.69                             & 1.19                          & 83.18  \\
\multicolumn{1}{|c|}{}                      & \multicolumn{1}{c|}{}                                 & \multicolumn{1}{c|}{Threshold} & 6.40                     & 2.51                       & 7.41                           & 0.30                            & 0.42                             & 0.73                          & 88.59  \\ \hline
\multicolumn{1}{|c|}{\multirow{2}{*}{fc2}}  & \multicolumn{1}{c|}{\multirow{2}{*}{Fully Connected}} & \multicolumn{1}{c|}{Bias}      & 3.75                     & 2.70                       & 8.45                           & 0.09                            & 0.39                             & 0.62                          & 87.71  \\
\multicolumn{1}{|c|}{}                      & \multicolumn{1}{c|}{}                                 & \multicolumn{1}{c|}{Weight}    & 3.75                     & 3.10                       & 3.95                           & 0.06                            & 0.18                             & 0.28                          & 92.40  \\ \hline
\multicolumn{1}{|c|}{\multirow{4}{*}{lif2}} & \multicolumn{1}{c|}{\multirow{4}{*}{Leaky}}           & \multicolumn{1}{c|}{Beta}      & 5.62                     & 3.08                       & 10.11                          & 0.08                            & 0.46                             & 0.71                          & 85.54  \\
\multicolumn{1}{|c|}{}                      & \multicolumn{1}{c|}{}                                 & \multicolumn{1}{c|}{Potential} & 2.81                     & \multicolumn{1}{c}{2.44}   & 12.30                          & 0.04                            & 0.06                             & 0.09                          & 85.05  \\
\multicolumn{1}{|c|}{}                      & \multicolumn{1}{c|}{}                                 & \multicolumn{1}{c|}{Spike}     & 3.75                     & \multicolumn{1}{c}{8.78}   & 10.54                          & 0.13                            & 0.21                             & 0.36                          & 79.03  \\
\multicolumn{1}{|c|}{}                      & \multicolumn{1}{c|}{}                                 & \multicolumn{1}{c|}{Threshold} & 4.68                     & \multicolumn{1}{c}{22.91}  & 5.55                           & 0.09                            & 0.33                             & 0.55                          & 70.54  \\ \hline

\hline
\multicolumn{10}{|c|}{SHD}                                                                                                                                                                                                                                                                                                               \\ \hline
\multicolumn{1}{|c|}{Layer Name}            & \multicolumn{1}{c|}{Layer Type}                       & \multicolumn{1}{c|}{Parameter} & \multicolumn{1}{c|}{n\%} & \multicolumn{1}{c|}{SDC 1} & \multicolumn{1}{c|}{SDC 0-5\%} & \multicolumn{1}{c|}{SDC 5-10\%} & \multicolumn{1}{c|}{SDC 10-20\%} & \multicolumn{1}{c|}{SDC 20\%} & Masked \\ \hline
\multicolumn{1}{|c|}{\multirow{2}{*}{fc1}}  & \multicolumn{1}{c|}{\multirow{2}{*}{Fully Connected}} & \multicolumn{1}{c|}{Bias}      & 0.31                     & 2.76                       & 16.84                          & 1.66                            & 1.77                             & 1.74                          & 75.21  \\
\multicolumn{1}{|c|}{}                      & \multicolumn{1}{c|}{}                                 & \multicolumn{1}{c|}{Weight}    & 4.46                     & 0.40                       & 5.47                           & 0.36                            & 0.31                             & 0.31                          & 93.12  \\ \hline
\multicolumn{1}{|c|}{\multirow{2}{*}{fb1}}  & \multicolumn{1}{c|}{\multirow{2}{*}{Fully Connected}} & \multicolumn{1}{c|}{Bias}      & 0.35                     & 1.87                       & 20.26                          & 2.25                            & 2.16                             & 1.96                          & 71.47  \\
\multicolumn{1}{|c|}{}                      & \multicolumn{1}{c|}{}                                 & \multicolumn{1}{c|}{Wegiht}    & 0.27                     & 0.55                       & 8.28                           & 0.59                            & 0.50                             & 0.47                          & 89.58  \\ \hline
\multicolumn{1}{|c|}{\multirow{4}{*}{lif1}} & \multicolumn{1}{c|}{\multirow{4}{*}{Leaky}}           & \multicolumn{1}{c|}{Beta}      & 0.37                     & 2.06                       & 32.58                          & 3.04                            & 2.91                             & 2.51                          & 56.88  \\
\multicolumn{1}{|c|}{}                      & \multicolumn{1}{c|}{}                                 & \multicolumn{1}{c|}{Potential} & 0.31                     & 3.17                       & 40.59                          & 2.31                            & 1.96                             & 2.18                          & 49.76  \\
\multicolumn{1}{|c|}{}                      & \multicolumn{1}{c|}{}                                 & \multicolumn{1}{c|}{Spike}     & 0.28                     & 13.59                      & 37.97                          & 2.82                            & 2.61                             & 3.14                          & 39.56  \\
\multicolumn{1}{|c|}{}                      & \multicolumn{1}{c|}{}                                 & \multicolumn{1}{c|}{Threshold} & 0.35                     & 3.60                       & 24.32                          & 2.28                            & 2.16                             & 2.95                          & 64.66  \\ \hline
\multicolumn{1}{|c|}{\multirow{2}{*}{fc2}}  & \multicolumn{1}{c|}{\multirow{2}{*}{Fully Connected}} & \multicolumn{1}{c|}{Bias}      & 0.31                     & 1.26                       & 53.56                          & 1.14                            & 0.90                             & 1.34                          & 41.77  \\
\multicolumn{1}{|c|}{}                      & \multicolumn{1}{c|}{}                                 & \multicolumn{1}{c|}{Weight}    & 0.28                     & 1.90                       & 16.74                          & 0.22                            & 0.18                             & 0.23                          & 80.67  \\ \hline
\multicolumn{1}{|c|}{\multirow{4}{*}{lif2}} & \multicolumn{1}{c|}{\multirow{4}{*}{Leaky}}           & \multicolumn{1}{c|}{Beta}      & 0.15                     & 0                          & 14.79                          & 0                               & 0                                & 0                             & 85.20  \\
\multicolumn{1}{|c|}{}                      & \multicolumn{1}{c|}{}                                 & \multicolumn{1}{c|}{Potential} & 0.15                     & 0                          & 0                              & 0                               & 0                                & 0                             & 100    \\
\multicolumn{1}{|c|}{}                      & \multicolumn{1}{c|}{}                                 & \multicolumn{1}{c|}{Spike}     & 0.31                     & 0                          & 0                              & 0                               & 0                                & 0                             & 100    \\
\multicolumn{1}{|c|}{}                      & \multicolumn{1}{c|}{}                                 & \multicolumn{1}{c|}{Threshold} & 0.62                     & 1.07                       & 26.39                          & 0.50                            & 0.43                             & 0.97                          & 70.61  \\ \hline

\hline
\multicolumn{10}{|c|}{DVS128}                                                                                                                                                                                                                                                                                                                \\ \hline
\multicolumn{1}{|c|}{Layer Name}             & \multicolumn{1}{c|}{Layer Type}                       & \multicolumn{1}{c|}{Parameter} & \multicolumn{1}{c|}{n\%} & \multicolumn{1}{c|}{SDC 1} & \multicolumn{1}{c|}{SDC 0-5\%} & \multicolumn{1}{c|}{SDC 5-10\%} & \multicolumn{1}{c|}{SDC 10-20\%} & \multicolumn{1}{c|}{SDC 20\%} & Masked \\ \hline
\multicolumn{1}{|c|}{\multirow{2}{*}{conv1}} & \multicolumn{1}{c|}{\multirow{2}{*}{Convolutional}}   & \multicolumn{1}{c|}{Bias}      & 2.86                     & 1.46                       & 13.46                          & 1.38                            & 1.49                             & 1.67                          & 80.54  \\
\multicolumn{1}{|c|}{}                       & \multicolumn{1}{c|}{}                                 & \multicolumn{1}{c|}{Weight}    & 1.98                     & 1.70                       & 23.04                          & 1.75                            & 1.61                             & 2.12                          & 69.78  \\ \hline
\multicolumn{1}{|c|}{\multirow{4}{*}{lif1}}  & \multicolumn{1}{c|}{\multirow{4}{*}{Leaky}}           & \multicolumn{1}{c|}{Beta}      & 1.56                     & 1.95                       & 25.95                          & 1.79                            & 2.06                             & 2.12                          & 66.13  \\
\multicolumn{1}{|c|}{}                       & \multicolumn{1}{c|}{}                                 & \multicolumn{1}{c|}{Potential} & 1.94                     & 1.44                       & 7.33                           & 0.52                            & 0.50                             & 0.71                          & 89.50  \\
\multicolumn{1}{|c|}{}                       & \multicolumn{1}{c|}{}                                 & \multicolumn{1}{c|}{Spike}     & 2.00                     & 1.59                       & 7.06                           & 0.51                            & 0.48                             & 0.69                          & 89.66  \\
\multicolumn{1}{|c|}{}                       & \multicolumn{1}{c|}{}                                 & \multicolumn{1}{c|}{Threshold} & 2.01                     & 4.90                       & 22.05                          & 1.82                            & 2.73                             & 4.38                          & 64.11  \\ \hline
\multicolumn{1}{|c|}{\multirow{2}{*}{conv2}} & \multicolumn{1}{c|}{\multirow{2}{*}{Convolutional}}   & \multicolumn{1}{c|}{Bias}      & 1.27                     & 5.41                       & 4.21                           & 0.33                            & 0.66                             & 0.72                          & 88.67  \\
\multicolumn{1}{|c|}{}                       & \multicolumn{1}{c|}{}                                 & \multicolumn{1}{c|}{Weight}    & 2.02                     & 0.73                       & 3.37                           & 0.25                            & 0.25                             & 0.36                          & 95.05  \\ \hline
\multicolumn{1}{|c|}{\multirow{4}{*}{lif2}}  & \multicolumn{1}{c|}{\multirow{4}{*}{Leaky}}           & \multicolumn{1}{c|}{Beta}      & 3.13                     & 1.48                       & 15.16                          & 1.25                            & 1.72                             & 1.80                          & 78.59  \\
\multicolumn{1}{|c|}{}                       & \multicolumn{1}{c|}{}                                 & \multicolumn{1}{c|}{Potential} & 1.91                     & 0.88                       & 5.05                           & 0.56                            & 0.57                             & 0.87                          & 92.06  \\
\multicolumn{1}{|c|}{}                       & \multicolumn{1}{c|}{}                                 & \multicolumn{1}{c|}{Spike}     & 2.05                     & 0.56                       & 3.66                           & 0.36                            & 0.40                             & 0.53                          & 94.48  \\
\multicolumn{1}{|c|}{}                       & \multicolumn{1}{c|}{}                                 & \multicolumn{1}{c|}{Threshold} & 2.50                     & 15.82                      & 22.27                          & 2.54                            & 2.05                             & 5.57                          & 51.76  \\ \hline
\multicolumn{1}{|c|}{\multirow{2}{*}{fc1}}   & \multicolumn{1}{c|}{\multirow{2}{*}{Fully Connected}} & \multicolumn{1}{c|}{Bias}      & 1.42                     & 0.00                       & 0.55                           & 0.00                            & 0.08                             & 0.08                          & 99.30  \\
\multicolumn{1}{|c|}{}                       & \multicolumn{1}{c|}{}                                 & \multicolumn{1}{c|}{Weight}    & 2.01                     & 0.44                       & 1.99                           & 0.07                            & 0.07                             & 0.10                          & 97.32  \\ \hline
\multicolumn{1}{|c|}{\multirow{4}{*}{lif3}}  & \multicolumn{1}{c|}{\multirow{4}{*}{Leaky}}           & \multicolumn{1}{c|}{Beta}      & 0.85                     & 6.12                       & 52.60                          & 2.99                            & 2.60                             & 2.86                          & 32.81  \\
\multicolumn{1}{|c|}{}                       & \multicolumn{1}{c|}{}                                 & \multicolumn{1}{c|}{Potential} & 2.27                     & 14.26                      & 62.94                          & 0.78                            & 0.98                             & 1.32                          & 18.41  \\
\multicolumn{1}{|c|}{}                       & \multicolumn{1}{c|}{}                                 & \multicolumn{1}{c|}{Spike}     & 2.27                     & 4.20                       & 62.79                          & 0.29                            & 1.22                             & 0.59                          & 30.91  \\
\multicolumn{1}{|c|}{}                       & \multicolumn{1}{c|}{}                                 & \multicolumn{1}{c|}{Threshold} & 1.14                     & 90.53                      & 4.10                           & 1.37                            & 1.86                             & 2.15                          & 0.00   \\ \hline

\end{tabular}
\end{table*}

Analyzing the results obtained using the simple \gls{FC} architecture applied to the N-MNIST dataset, one noticeable trend is that as the injection is directed towards layers closer to the output, the impact becomes increasingly pronounced. This observation is evident in the diminishing number of masked faults with increasing layer index and the different categories of \gls{SDC}, which exhibit an opposite trend.
Zooming in on individual parameters, the impact on the bias in linear layers is more significant than that on synaptic weights. This is reasonable because, typically, a synaptic weight represents the contribution of a single input to the total synaptic current. In contrast, the bias influences the overall value of the current itself, often having a larger magnitude and, consequently, a stronger impact.
Looking at the spiking layers then, central to this work, the impact of a fault in each of the model parameters can be seen. In particular, faults affecting the threshold or the output spike have the highest impact. This was again expected since they directly affect the neuron activity, leading in extreme cases to the aforementioned dead and saturated neuron situations, which are the harsher ones.
Results on the recurrent \gls{FC} model used on the \gls{SHD} are in some ways comparable to the N-MNIST ones. The impact of the faults, in general, increases moving towards the outer layers. Interestingly, faults in the feedback connections cause more errors than the \gls{FF} ones. Results on the last \gls{LIF} layer, particularly the membrane potential and spike, are probably outliers and need further exploration, targeting them with specific fault injection campaigns.
Finally, the \gls{CSNN} applied to the DVS128 dataset has a quite different behavior: faults in the first convolutional layers have a larger impact from the point of view of \glspl{SDC}, indicating that the initial feature extraction phase is likely the most crucial aspect in such architectures.
On the other hand, \gls{LIF} layers behave in the usual way: faults impact more in outer layers. This is even more evident concerning the previous models, suggesting that the phenomenon is even stronger with deeper architecture. Additionally, the model time constant is the second most relevant quantity, immediately after the threshold. Given the temporal relevance of features in gesture recognition, this underlines the importance of temporal information in this specific model, meaning that timing variations in the neurons' activity, i.e., byzantine neuron behaviors, are more impactful in this case.
%
%
However, it has to be noticed that the presented results are preliminary, and a more thorough and comprehensive analysis should be conducted, focusing on injecting faults into individual layers. Indeed, the network-wise fault injection may cause the error margin at the layer level to be greater than the predefined value ($e=1\%$). Consequently, results can hint at the general reliability of different network parameters.
Nonetheless, this initial analysis already showcases the capabilities of \spikingjet{} in conducting comprehensive fault injection campaigns within reasonable timeframes. Moreover, the results of the analysis highlight the resilience of different families of \gls{SNN} models against stuck-at faults.
\section{Conclusions}
\label{sec:conclusions}

In conclusion, this paper introduces \spikingjet{}, a \gls{FI} tool to assess the resilience of \gls{FC} and convolutional \glspl{SNN} to hardware faults. \spikingjet{} provides a comprehensive platform for assessing the resilience of \glspl{SNN} by inducing errors and injecting faults into critical components such as synaptic weights, neuron model parameters, internal states, and activation functions. A first analysis conducted with \spikingjet{} shows the overall robustness of \glspl{SNN} to stuck-at faults. Simultaneously, it enables the identification of the network's weakest components, shedding light on areas where focusing fault tolerance techniques would be beneficial.
\spikingjet{} establishes a foundation for in-depth examination of \gls{SNN} resilience, starting with a comprehensive analysis of the network and subsequently concentrating on its most vulnerable components. With its rapid experimental pace, it facilitates agile and adaptable evaluation of \gls{SNN} resilience.


\AtNextBibliography{\small}
\printbibliography


\end{document}